\documentclass[letterpaper]{article} % DO NOT CHANGE THIS
\usepackage{aaai24}  % DO NOT CHANGE THIS
\usepackage{times}  % DO NOT CHANGE THIS
\usepackage{helvet}  % DO NOT CHANGE THIS
\usepackage{courier}  % DO NOT CHANGE THIS
\usepackage[hyphens]{url}  % DO NOT CHANGE THIS
\usepackage{graphicx} % DO NOT CHANGE THIS
\usepackage{natbib}  % DO NOT CHANGE THIS AND DO NOT ADD ANY OPTIONS TO IT
\usepackage{caption} % DO NOT CHANGE THIS AND DO NOT ADD ANY OPTIONS TO IT
\usepackage{algorithm}
\usepackage{algorithmic}

\usepackage{newfloat}
\usepackage{listings}
\DeclareCaptionStyle{ruled}{labelfont=normalfont,labelsep=colon,strut=off} % DO NOT CHANGE THIS
\floatstyle{ruled}
\newfloat{listing}{tb}{lst}{}
\floatname{listing}{Listing}
\title{Dynamic Strategy Chain: Dynamic Zero-Shot CoT \\for Long Mental Health Support Generation}
\author{
    Qi Chen\equalcontrib,
    Dexi Liu\equalcontrib
}
\affiliations{
    \textsuperscript{\rm 1}Jiangxi University of Finance and Economics\\
}

\usepackage{bibentry}
\begin{document}

\maketitle

\begin{abstract}
Long counseling Text Generation for Mental health support (LTGM), an innovative and challenging task, aims to provide help-seekers with mental health support through a comprehensive and more acceptable response. The combination of chain-of-thought (CoT) prompting and Large Language Models (LLMs) is employed and get the SOTA performance on various NLP tasks, especially on text generation tasks. Zero-shot CoT prompting is one of the most common methods in CoT prompting. However, in the LTGM task, Zero-shot CoT prompting can not simulate a counselor or provide personalized strategies without effective mental health counseling strategy prompts. To tackle this challenge, we propose a zero-shot Dynamic Strategy Chain (DSC) prompting method. Firstly, we utilize GPT2 to learn the responses written by mental health counselors and dynamically generate mental health counseling strategies tailored to the help-seekers needs. Secondly, the Zero-shot DSC prompting is constructed according to mental health counseling strategies and the help-seeker’s post. Finally, the Zero-shot DSC prompting is employed to guide LLMs in generating more human-like responses for the help-seekers. Both automatic and manual evaluations demonstrate that Zero-shot DSC prompting can deliver more human-like responses than CoT prompting methods on LTGM tasks.
\end{abstract}

\section{Introduction}

According to the latest data from the World Health Organization, there are approximately 450 million individuals worldwide with mental health disorders \cite{bib1}. Mental health issues are becoming increasingly severe, causing immense pain to individual lives and affecting the overall health and well-being of society \cite{bib2,bib3}. Online mental health counseling, as an effective therapy for mental disorders \cite{bib4}, has become popular in recent years (Sun et al. 2021).

In recent years, significant advancements have occurred in NLP and AI, primarily due to the emergence of large language models (LLMs). Noteworthy models such as GPT-3 \cite{bib5}, PaLM \cite{bib6}, Llama \cite{bib7} and GPT-3.5 (OpenAI, 2023) have demonstrated the potential of LLMs by leveraging their increasing model sizes and vast amounts of training data. As a result, these models have achieved human-level performance across various tasks, including summarization, translation, question answering, and basic mathematical reasoning \cite{bib8}. Figure 1 shows the results generated by GPT3.5-turbo in the Long counseling Text Generation for Mental health support (LTGM) task.

\begin{figure}[t]
\centering
\centerline{\includegraphics[scale=0.35]{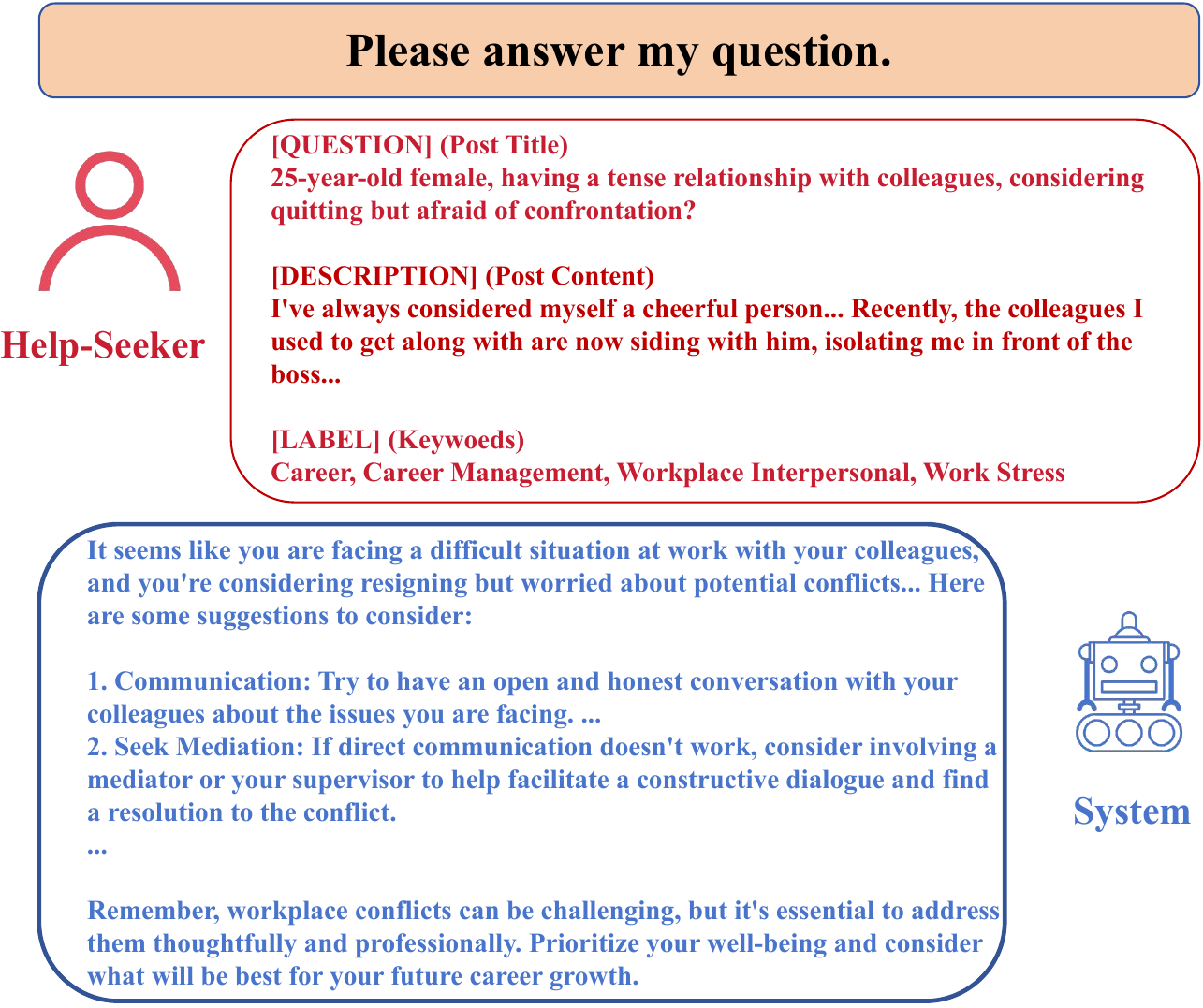}}%%图像路径：pic是文件夹名。
\caption{An example showing Long counseling Text Generation for Mental Health support, which includes a system(GPT3.5-turbo) for seeking helpers and providing mental health support services. The prompt 'Please answer my question.' is used to guide the system to respond to the Help-Seeker's question.}
\label{ov}
\end{figure}

Despite demonstrating strong coherence and structural awareness in generating lengthy text, LLMs have certain limitations in the LTGM task, as illustrated in Figure 1. Firstly, when answering mental health support questions, LLMs skip the expression of understanding and analysis, providing generic advice that often comes across as indifferent and unresponsive. Secondly, it lacks guided strategies for target emotions, making it challenging to effectively drive users out of emotional distress. Thirdly, LLMs rarely generate personalized responses based on different users' language styles, needs, and preferences, resulting in answers lacking individualization and customization. These limitations constrain the performance of LLMs in mental health support conversation scenarios and call for further improvements.

\citet{bib9} provided a classification of language assistance skills and strategies for mental health counselors. In Hill's theory, language assistance skills for mental health counselors are classified into various strategies. \citet{bib10} referred to Hill's theory and categorized mental health support skills into eight categories, each with its corresponding strategy and definition, as presented in Table 1. We reference the strategy collection by Liu et al. (2021) and find that using strategies can guide LLMs to respond to questions from different perspectives. For example, when given the strategy "[Self-disclosure]," LLMs search its database for similar experiences to the helper's question and respond. Below is an example of LLMs generating a response based on the "[Self-disclosure]" strategy: "I have also faced a similar problem in the past, and I felt confused and distressed at that time. However, I sought help from my mentors and peers, and they provided me with valuable advice and assistance, helping me overcome the challenge. I believe you can also resolve your issue by communicating with those around you."

\begin{table}[t]
    \centering
    \renewcommand{\arraystretch}{1.5} % 增加行高
    \begin{tabular}{p{3cm}p{5cm}}
        \hline
        \textbf{Strategy} & \textbf{Strategy Description} \\
     \hline
        Information & Providing data, facts, opinions, and resources as information \\
   
        Direct Guidance & Offering advice, instructions, directions, or suggestions on what the help-seeker should do to bring about change \\
  
        Approval And Reassurance & Giving mental health support, comfort, encouragement, and reinforcement \\
   
        Restatement & Simply repeating or restating the content or meaning of the problem, usually in a more specific and clear way \\

        Interpretation & Going beyond what the help-seeker has already expressed or recognized, providing new meanings, reasons, or explanations \\

        Self-disclosure & Revealing personal information about the helper's non-direct experiences or feelings \\
        \hline
    \end{tabular}
    \caption{Strategy Collection in Counseling}
    \label{table1}
\end{table}

\begin{figure}[t]
\centering
\centerline{\includegraphics[scale=0.30]{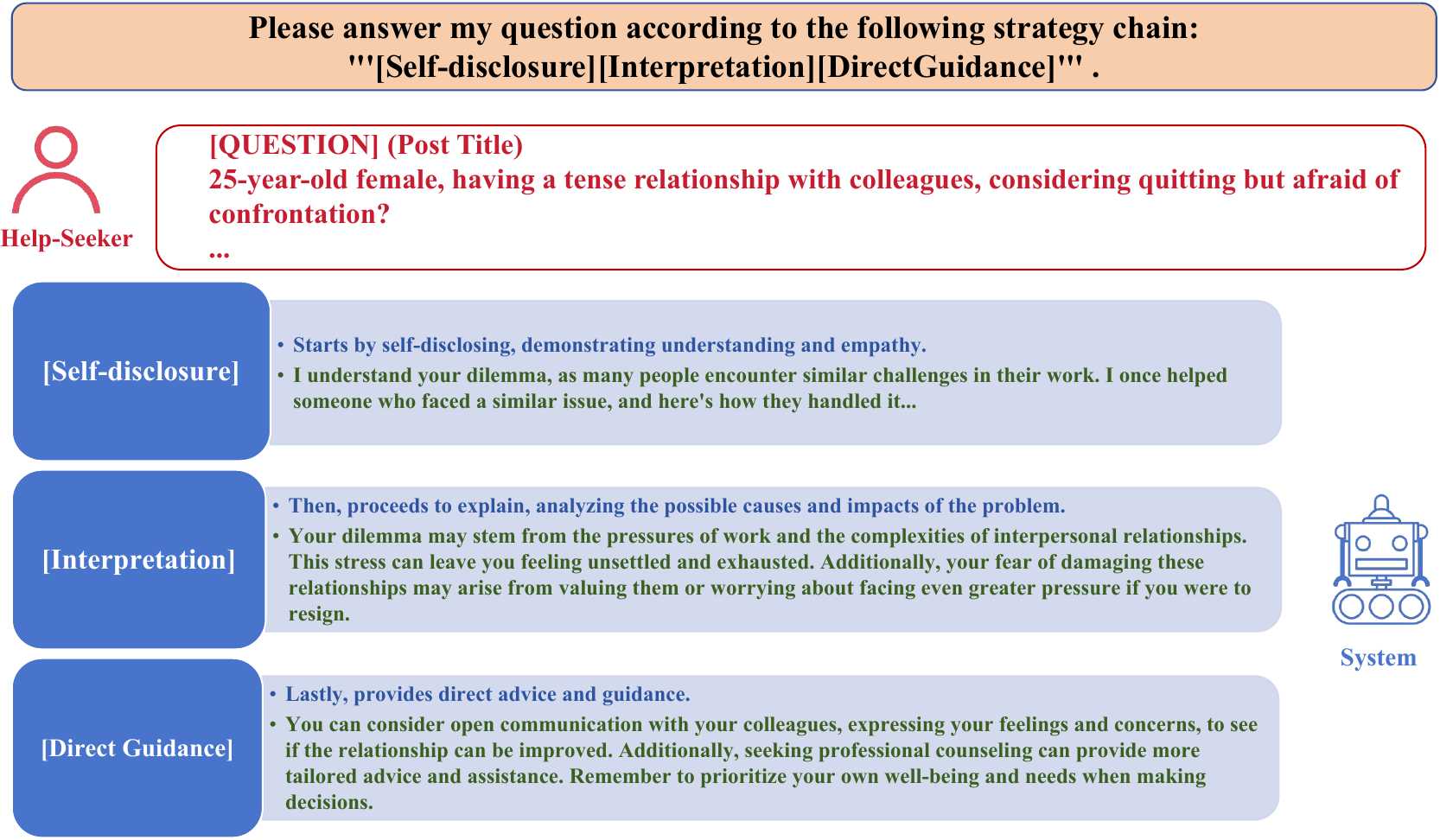}}%%图像路径：pic是文件夹名。
\caption{An example shows the generated results of the Zero-shot DSC prompting method. Black font indicates a prompting. The red font indicates a helper question. The white font indicates the DSC. The blue font indicates the GPT3.5-turbo interpretation of the DSC. The green font indicates the final generated result of GPT3.5-turbo.}
\label{ov}
\end{figure}

Inspired by this, We utilize support strategies to overcome the limitations of LLMs in LTGM tasks. We propose the concept of a Dynamic Strategy Chain (DSC), which is the sequence of strategies dynamically generated based on the queries of those seeking mental health assistance. The DSC is a guiding framework for the system to generate comprehensive mental health support responses. Compared to the Zero-shot chain-of-thought (CoT)\cite{bib15} prompting of "Think it step by step," Zero-shot DSC prompting can provide more specific and dynamic guidance. For instance, for the question, "A 25-year-old woman, having a tense relationship with colleagues, wants to quit but fears breaking ties", we can generate a DSC: [Self-disclosure][Interpretation][Direct Guidance]. Figure 2 shows the results of generating mental health support responses for the Zero-shot DSC prompting method. Prompted by the DSC, the LLMs will first share feelings or experiences, then analyze the problem, and finally provide advice. The core idea of this method is to use a specialized mental health domain model to learn domain knowledge, generate the DSC, and then apply DSC in the dialogue generation process to achieve dynamic guidance.

In this paper, we employ GPT2 to generate multiple DSCs for help-seeker's questions, and then prompt LLMs to generate the most suitable personalized responses. Our method enables the LLMs to adapt more effectively to various situations and user requirements, resulting in the generation of more personalized and targeted responses.
Contributions. We summarize our contributions and innovations as follows:
\begin{itemize}
\item This paper represents the first research work on the application of LLMs in the field of mental health.

\item This paper is the first work to use Pre-trained Language Models (PLMs) to guide LLMs in generating long texts.

\item This paper is the first work to propose the concept of dynamic prompts.
\end{itemize}

\begin{figure*}[t]
\centering
\centerline{\includegraphics[scale=0.9]{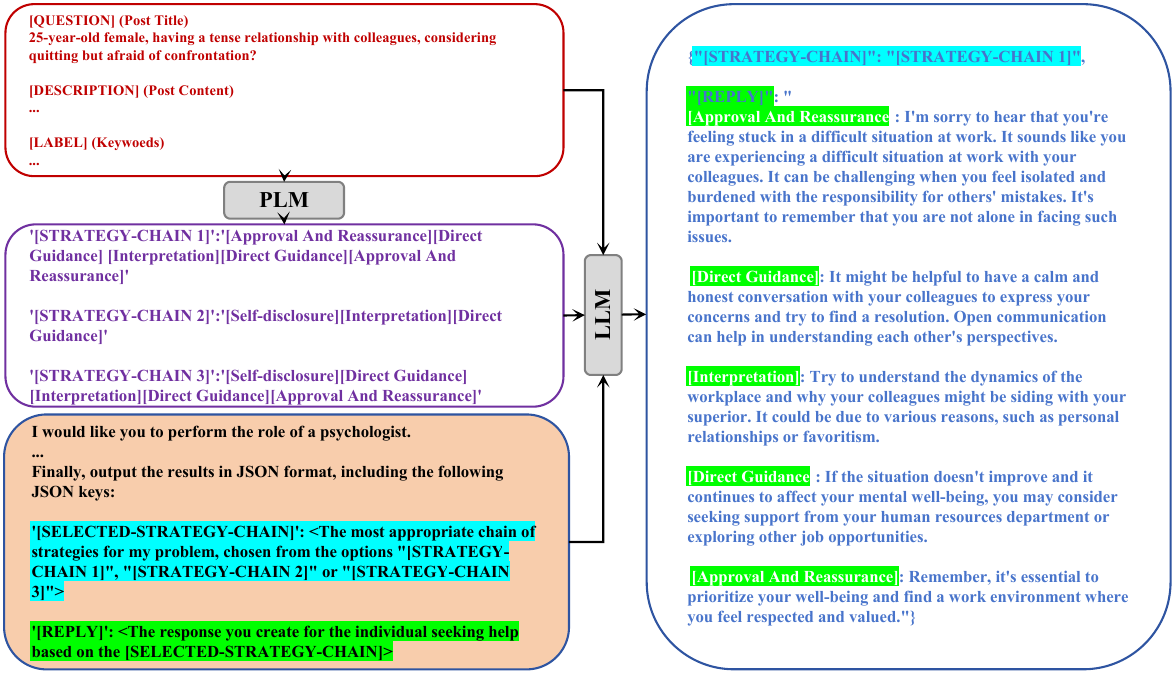}}%%图像路径：pic是文件夹名。
\caption{Examples of Zero-shot DSC prompting. Purple font indicates DSC generated by GPT2. The sentences with a blue background indicate the prompting that guided the LLMs' selection, while the sentences with a green background represent the prompting generated based on the guidance of the LLMs. On the right, we have the final generated results. The blue and green backgrounds on the right correspond to the blue and green backgrounds on the left. The meaning of the other colored fonts is the same as that in Figure 2.}
\label{ov}
\end{figure*}

\section{Related Work}
\textbf{Emotional Support.} Emotional Support Conversations have received much attention in recent years. Liu et al. (2021) proposed the Emotional Support Conversation task and released the dataset ESConv(Emotional Support Conversation dataset). They attach the support strategy as a special token at the beginning of each sentence to prompt the generation of that sentence. Similarly, Sun et al. (2021) published a high-quality Chinese mental health support dataset (PsyQA) based on the support strategy. Currently, in emotional support tasks, most experimental models rely on Pre-trained Language Models (PLMs), with researchers focusing on effective architectures to integrate external knowledge for generating more informative, diverse, and generative responses \cite{bib11,bib12,bib13}. For instance, Peng et al.(2022 ) proposed a hierarchical graph network to simultaneously exploit global emotional reasons and local user intentions generated by COMET \cite{bib14}. Tu et al. (2022) incorporated COMET-generated common-sense knowledge and a hybrid response strategy into emotionally-supportive dialogues instead of using a single strategy to generate responses. Li et al. (2022) and others considered how to leverage external knowledge bases to improve empathy in the context of empathic emotional understanding and expression in dialogue generation tasks.

In summary, Emotional Support Conversation tasks involve a significant amount of common knowledge and knowledge from various domains, with a focus on utilizing external knowledge through PLMs. However, when compared to LLMs, PLMs have limited capabilities in acquiring knowledge. Even with the supplementation of external knowledge, the parameter capacity of PLM remains significantly lower than that of LLMs.

\textbf{Prompting Methods.} Prompting (OpenAI 2023) is an innovation in the field of LLMs that aims to guide LLMs to generate more flexible and efficient outputs by providing specific inputs or hints. Recently, there have been many researches based on Prompting, \citet{bib15} proposed thought Zero-shot CoT prompting to enhance the performance of LLMs in complex reasoning problems by adding hints “Let's think step by step.” before the answer to the question. \citet{bib16}introduced the Zero-shot Plan and Solved (PS) Prompting, which involves dividing tasks into subtasks and executing them according to a devised plan. However, both Zero-shot CoT prompting and Zero-shot PS prompting are static prompting, which means they cannot be adjusted as the problem changes \citet{bib17} proposed the Zero-shot Tree of Thought (ToT) prompting, which is a dynamic cue that can dynamically change its strategy with the problem. It allows language models to be explored and evaluated at each step of the problem-solving process to obtain better decision-making options. 

However, despite Zero-shot ToT prompting's superior performance in tasks such as complex problem reasoning, the disregard for global information also imposes constraints on its effectiveness in generating long texts. For this reason, we propose the Dynamic and Low Resource Requirement Cueing approach, Zero-shot DSC prompting. It utilizes PLMs to generate multiple DSCs. Subsequently, LLMs are employed to select the most suitable DSC based on the given problem. Compared to Zero-shot ToT prompting, the Zero-shot DSC prompting method reduces the number of queries to one, which significantly alleviates the resource requirements for LLMs querying. Additionally, the Zero-shot DSC prompting method provides multiple global strategies, offering comprehensive guidance for LLMs' global thinking. This approach simultaneously leverages the advantages of PLMs, which are fine-tunable and easy to train on specific tasks, and LLMs, which excel in problem analysis, reasoning, fluent response generation, and encompassing a wealth of common knowledge.

\begin{figure*}[h]
\centering
\centerline{\includegraphics[scale=0.45]{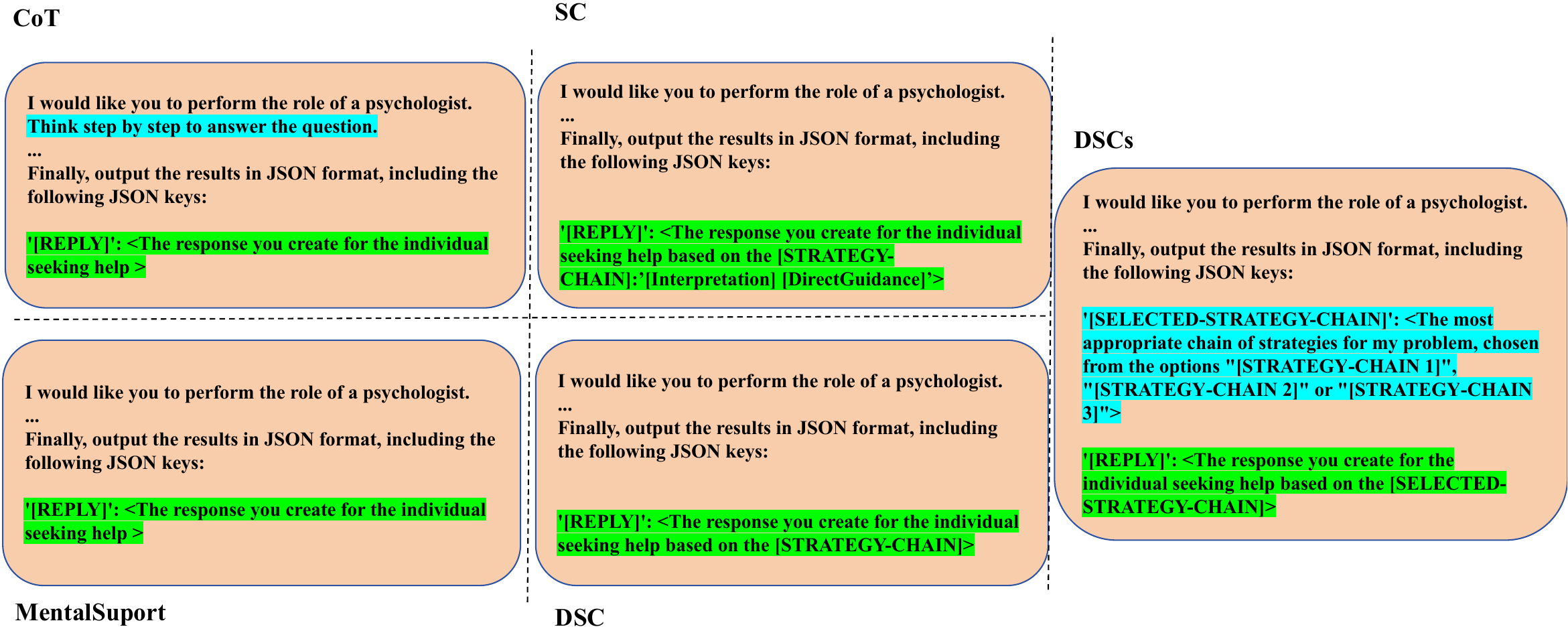}}%%图像路径：pic是文件夹名。
\caption{ Prompting of methods (2) to (6).}
\label{ov}
\end{figure*}

\section{Methodology}

\textbf{Step 1: Guiding PLMs to Generate DSC}
To provide references and guidance for LLMs in generating responses, we learn the dialogue Strategy Chain (SC) from mental health counselors.

We follow the approach of Sun et al. (2021) by adding the strategy as a special token at the beginning of each segment and using cross-entropy loss as the loss function. Specifically, the format of the model input is [QUESTION] Q [DESCRIPTION] D [LABEL] L [STRATEGY-CHAIN] DSC, where Q, D, L, and DSC are separated by predefined special tokens. Similarly, the text representation for DSC can be expressed as
[Strategy1][Strategy2][Strategy3]...
As shown in Figure 3, we generate three strategy chains at once: "[STRATEGY-CHAIN 1]", "[STRATEGY-CHAIN 2]", or "[STRATEGY-CHAIN 3]".

\textbf{Step 2: Guiding LLMs in Response Generation}
LLMs often have fixed thinking patterns and tend to have shallow thinking. To provide personalized responses and encourage deeper thinking, we offer dynamic and diverse global prompts. These global prompts require LLMs to consider the strategy chain from a holistic and personalized perspective for a better understanding of the content to be answered. For example, we first select the most suitable strategy chain from "[STRATEGY-CHAIN 1]", "[STRATEGY-CHAIN 2]", or "[STRATEGY-CHAIN 3]" which are generated from step 1, and then generate a response based on the selected strategy chain. 
As shown in Figure 3, the prompt would be

'[SELECTED-STRATEGY-CHAIN]': \textless The most appropriate chain of
strategies for my problem, chosen from the options "[STRATEGY- CHAIN 1]", "[STRATEGY-CHAIN 2]" or "[STRATEGY-CHAIN 3]"\textgreater

'[REPLY]': \textless The response you create for the individual seeking help
based on the [SELECTED-STRATEGY-CHAIN]\textgreater

Contrary to Zero-shot CoT prompting, which encourages step-by-step thinking, Zero-shot DSC prompting provides more specific strategic plans. Zero-shot DSC prompting enables LLMs to have a global dynamic plan when answering questions, considering the interaction between the entire question and response in advance. This approach allows for the generation of more personalized and targeted responses.

\begin{table*}[htbp]
    \centering
    \renewcommand{\arraystretch}{1.2} % 增加行高
    \begin{tabular}{lccccccc}
    \hline
    Model          & Blue 1 & Blue 2 & Blue 3 & Blue 4 & Blue Avg & D1   & D2    \\ \hline
    GPT$_\mathrm{ft}$+strategy  & \textbf{48.96}  & \textbf{25.55}  & \underline{11.26}  & \underline{4.76}   & \textbf{22.63}    & 7.36 & \textbf{40.47} \\
    CoT            & 43.36  & 21.99  & 9.84   & 4.37   & 19.89    & 8.51 & 36.63 \\
    \hline
    MHS  & 42.27  & 21.95  & 10.09  & 4.55   & 19.71    & \textbf{8.91} & 37.28 \\ 
    SC  & 41.24  & 20.97  & 9.61   & 4.34   & 19.04    & 8.64 & 36.46 \\ 
    DSC  & 43.21  & 21.97  & 9.84   & 4.33   & 19.83    & \underline{8.85} & \underline{37.58} \\ 
    DSCs    & \underline{46.26}  & \underline{24.05}  & \textbf{11.34}  & \textbf{5.42}   & \underline{21.77}    & 8.74 & 37.57 \\ 
     \hline
    \end{tabular}
    \caption{Automatic evaluation results (\%). Bold numbers indicate the highest value for that metric, while underlined numbers indicate the second-highest value for that metric.}
    \label{table1}
\end{table*}

\section{Dataset}

In order to assess the effectiveness of the Zero-shot DSC prompting method, an experimental study was conducted using the PsyQA dataset(Sun et al. 2021), a high-quality Chinese dataset that focuses on mental health counseling. This dataset comprises a total of 22,346 questions and 56,063 corresponding answers. On average, each question consists of 21.6 characters, each question description is around 168.9 characters long, and each answer contains approximately 524.6 characters. Within the dataset, 4,012 questions and answers were manually annotated with strategies, which accounts for about 17.9\% of the total number of questions and 7.1\% of the total number of answers. The remaining questions and answers were annotated automatically using the RoBERTa model. Each answer is annotated with an average of 6.66 strategies and 3.65 of them are unique. PsyQA covers a wide range of topics, including 9 broad subjects and multiple subtopics. 

To address challenges pose by excessively long strategy chains and closely repeated strategies, both of which can lead to confusion during generation by LLMs, a two-step data preprocess approach was employed. \textbf{The first step, strategy merging.} Involve the consolidation of adjacent identical strategies. For instance, a strategy chain like "[Information][Interpretation] [Interpretation][Interpretation][Interpretation][Information]" is condensed to "[Information] [Interpretation][Information]". This consolidation aim to streamline and simplify the strategy chains, making them more coherent. \textbf{The second step, sample filtering.} According to the statistical analysis of SC lengths in the data, it is found that long SC has a smaller proportion in the samples. Therefore, SC with a frequency lower than 5\% are removed, and only samples with SC lengths less than or equal to 8 are retained.

\section{Baselines}
We compare our model on the following classic generation structure: 
\textbf{(1) GPT$_\mathrm{ft}$+strategy (Sun et al. 2021):}
Uses GPT2 to generate replies with support strategies, aimed at providing mental health support and assistance.
\textbf{(2) CoT (Kojima et al. 2022):}
Zero-shot CoT prompting, appends "Think step by step to answer the question" to the prompt, guiding the generation of structured responses.
\textbf{(3) MHS:}
Provides a prompt labeled "Mental Health Support" for LLMs to generate responses.
\textbf{(4) SC:}
Extracts the most frequent strategy chain from the PsyQA dataset and uses it as a prompt, enabling LLMs to generate responses based on those strategies.
\textbf{(5) DSC:}
Offers a specific dominant strategy chain as a prompt, guiding LLMs to generate responses emphasizing particular strategies.
\textbf{(6) DSCs:}
Supplies multiple dominant strategy chains, allowing LLMs to choose the most appropriate one and generate responses accordingly.
Methods (2) to (6) are depicted in Figure 4.

\begin{table}[ht]
    \centering
    \small
    \tabcolsep=0.1cm
    \renewcommand{\arraystretch}{1.2} % 增加行高
    
    \begin{tabular}{lcccc}
    \hline
    Model & Fluency & Relevance & Helpfulness & Empathy \\
    \hline
    GPT$_\mathrm{ft}$+strategy & 3.55 & 3.10 & 2.49 & 2.70 \\
    CoT & \underline{4.78} & 4.51 & 3.62 & 3.17 \\
    Human & 4.62 & 4.53 & \underline{3.73} & 3.12 \\
    \hline

    MHS & 4.75 & 4.50 & 3.65 & 3.12 \\
    SC & 4.74 & 4.44 & 3.79 & 3.10 \\
    DSC & \textbf{4.81} & \underline{4.53} & 3.68 & \underline{3.25} \\
    DSCs & 4.76 & \textbf{4.65} & \textbf{3.86} & \textbf{3.54} \\

    \hline
    \end{tabular}

    \caption{Human evaluation results. Cohen’s Weighted Kappa \cite{bib20} score of 0.758 indicates a moderate level of agreement.}
    \label{table1}
\end{table}
\begin{table*}[t]
    \centering
    \small
    \tabcolsep=0.05cm

    \renewcommand{\arraystretch}{1.2} % 增加行高
    % \begin{adjustbox}{width=\linewidth}

   \begin{tabular}{ccccc|cccc|cccc}
    \hline
    \multicolumn{5}{c|}{Green} & \multicolumn{4}{c|}{Amber} & \multicolumn{4}{c}{Red} \\
    \hline
    Model & Fluency &  Relevance & Helpfulness & Empathy & Fluency &  Relevance & Helpfulness & Empathy & Fluency &  Relevance & Helpfulness & Empathy \\
    \hline
    GPT$_\mathrm{ft}$+strategy & 3.47 & 3.04 & 2.39 & 2.51 & 3.64 & 3.26 & 2.56 & 2.86 & 3.60 & 2.93 & 2.67 & 2.93 \\
    CoT & \textbf{4.82} & 4.54 & \underline{3.68} & 3.05 & 4.76 & 4.46 & 3.57 & 3.27 & \underline{4.70} & \underline{4.57} & 3.53 & 3.33 \\
    Human & 4.66 & 4.46 & 3.66 & 2.81 & 4.56 & \underline{4.56} & 3.74 & 3.33 & 4.63 & \textbf{4.67} & 3.90 & \underline{3.63} \\
    \hline

    MHS & 4.76 & 4.47 & 3.61 & 2.99 & 4.76 & 4.54 & 3.63 & 3.23 & \underline{4.70} & 4.50 & 3.80 & 3.27 \\
    SC & 4.75 & 4.41 & 3.67 & 2.94 & \underline{4.76} & 4.50 & \underline{3.90} & 3.27 & \underline{4.70} & 4.37 & \underline{3.93} & 3.23 \\
     DSC & \underline{4.80} & \underline{4.58} & 3.60 & \underline{3.13} & \textbf{4.87} & 4.54 & 3.80 & \underline{3.37} & \underline{4.70} & 4.33 & 3.67 & 3.33 \\
    DSCs & 4.78 & \textbf{4.65} & \textbf{3.72} & \textbf{3.33} & 4.71 & \textbf{4.66} & \textbf{4.00} & \textbf{3.71} & \textbf{4.80} & \underline{4.60} & \textbf{3.97} & \textbf{3.80} \\
   \hline
    \end{tabular}
    % \end{adjustbox}
    \caption{Human evaluation results of different mental health risk levels. }
    \label{table1}
\end{table*}

\begin{table*}[t]
    \centering
    \small
    \tabcolsep=0.05cm
    \renewcommand{\arraystretch}{1.2} % 增加行高
    % \begin{adjustbox}{width=\linewidth}
    \begin{tabular}{ccccc|cccc|cccc}
    \hline
    \multicolumn{5}{c|}{Simple} & \multicolumn{4}{c|}{Moderate} & \multicolumn{4}{c}{Complex}\\
    \hline
    Model & Fluency &  Relevance & Helpfulness & Empathy & Fluency &  Relevance & Helpfulness & Empathy & Fluency &  Relevance & Helpfulness & Empathy \\
    \hline
    GPT$_\mathrm{ft}$+strategy & 2.88 & 2.63 & 2.08 & 2.38 & 3.57 & 3.05 & 2.20 & 2.71 & 3.60 & 3.05 & 2.93 & 2.85 \\
    CoT & \underline{4.83} & \underline{4.83} & \underline{4.00} & 2.71 & \textbf{4.80} & 4.68 & 3.70 & 2.98 & \textbf{4.78} & 4.53 & 3.65 & 3.28 \\
    Human & 4.63 & 4.58 & 3.83 & 2.50 & 4.59 & 4.50 & 3.45 & \underline{3.02} & 4.57 & 4.45 & 3.78 & \underline{3.38} \\
    \hline
    
    MHS & 4.83 & 4.75 & 3.96 & 2.54 & 4.73 & 4.66 & 3.70 & 2.95 & 4.72 & \textbf{4.62} & 3.65 & 3.05 \\
    SC & \underline{4.83} & 4.54 & 3.92 & 2.58 & 4.70 & 4.66 & \textbf{3.89} & 2.89 & 4.60 & 4.50 & \underline{3.90} & 3.32 \\
    DSC & \textbf{4.96} & \textbf{4.88} & 3.96 & \textbf{3.21} & \textbf{4.80} & \textbf{4.80} & 3.73 & 2.93 & \underline{4.73} & 4.43 & 3.55 & 3.28 \\
     DSCs & 4.75 & \underline{4.83} & \textbf{4.17} & \underline{2.92} & \underline{4.77} & \underline{4.77} & \underline{3.84} & \textbf{3.39} & \underline{4.73} & \underline{4.60} & \textbf{3.95} & \textbf{3.68} \\
    
    \hline
    \end{tabular}
    % \end{adjustbox}
    \caption{Human evaluation results of question complexity. We judge the complexity of the question based on the number of topics it contains. We classify questions with 0-3 topics as simple questions, 4-6 topics as moderate questions, and 7 or more topics as complex questions.}
    \label{table1}
\end{table*}

\section{Evaluation Metrics}

\textbf{Automatic evaluation.}
The automatic metrics we adopted include BLEU \cite{bib18}, Distinct-1 (D1), and Distinct-2 (D2) \cite{bib19}. BLEU measures the similarity between the generated text and the responses of mental health counselors, while D1 and D2 measure the richness of vocabulary in the responses.

\textbf{Human evaluation.}
To evaluate the quality of the generated responses, we conducted a manual evaluation. We recruited 10 graduate students specializing in mental health or computer science to annotate the answers. These expert annotators were asked to rate the answers based on four criteria: Fluency, Relevance, Helpfulness, and Empathy(Liu et al. 2021).
\begin{itemize}
\item Fluency: Which bot's responses were more fluent and understandable?
\item Relevance: Which bot explored your situation more in-depth and was more helpful in identifying your problems?
\item Helpfulness: Which bot gave you more helpful suggestions for your problems?
\item Empathy: Did the bot show warmth, sympathy, and concern? Is there room for improvement?
\end{itemize}

The rating was done on a 5-star scale, where 5 stars represented the highest rating. 
To evaluate the quality of the generated responses, we followed the following steps:\\
(i) We randomly selected 100 questions as evaluation samples.\\
(ii) We divided the 10 graduate students specializing in mental health or computer science into five groups, with each group consisting of two annotators.\\
(iii) Each group's two annotators independently annotated 20 questions.\\
(iv) We calculated the score differences between the two annotators within each group for the same questions.\\
(v) We selected the answers with score differences exceeding two points and assigned them to another group of annotators for re-annotation.\\
(vi) We repeated steps (iv) and (v) until achieving a sufficiently high level of agreement (assessed using quadratic Cohen's Kappa).\\

\section{Evaluation Result}

\textbf{Automatic evaluation.} As show in Table 2, DSCs and GPT$_\mathrm{ft}$+strategy have the highest scores in terms of Blue value. Especially when the n-gram is improved, DSCs' advantage is more pronounced, surpassing GPT$_\mathrm{ft}$+strategy in the cases of Blue 3 and Blue 4. It is worth noting that without seeing any human responses during training, DSCs still achieve good results, indicating that the method proposed in this paper is effective in guiding LLMs to generate responses that are closer to human answers.

\textbf{Human evaluation.}The results presented in Table 3 show that the answers generated by LLMs guided by DSCs, through dynamic prompting and global thinking guidance, are preferred by human evaluators compared to other baseline methods. Our approach demonstrates significant advantages in terms of Relevance, Helpfulness, and Empathy, indicating that it excels in analyzing problems, providing suggestions, and comforting others. In comparison to human-generated responses, the answers generated by our method are more popular in all metrics, particularly in terms of Empathy. We speculate that this is because some human responses do not express Empathy adequately, as they tend to provide direct advice. Comparing DSCs with DSC, DSCs have higher Empathy values because they provide multiple choices for LLMs to select from. This not only reduces the impact of incorrect DSC generation but also guides LLMs in analyzing the given problem.

\textbf{DSC on different mental health risk levels.} To analyze the effectiveness of DSCs across different severity levels of mental disorders, we employed psychologists to annotate 100 extracted questions based on whether immediate mental health support help was required. 
Based on the risk levels referenced from CLPsych2017, we categorize posts into three levels: Red, Amber, and Green. Red: Posts that require immediate response and intervention for the expressed mental health issues. Amber: Posts that express mental health issues can be responded to at a later point. Green: Posts that can be safely ignored or left for the community to address.
The questions were categorized into three groups:
50 questions were classified as Green;
35 questions were classified as Amber;
15 questions were classified as Red.
The data from Table 4 revealed that the more severe the mental health problem, the higher the Empathy score from DSCs.
This suggests that DSCs can effectively adapt to and address a range of mental health issues, particularly those cases that require immediate mental health assistance. This also demonstrates the superiority and significance of DSCs in dealing with severe mental health problems.
Compared to the CoT, the proposed DSCs method consistently performed better in terms of Relevance, Helpfulness, and Empathy. This indicates that DSCs can generate not only more relevant and helpful responses but also better understand and empathize with the feelings of those seeking help, providing more human-centric and empathetic responses.

\textbf{DSC on different complexity of the question.} To examine the effectiveness of DSCs across varying levels of question complexity, we annotated these 100 questions with topics. Referring to the “Theory and Practice of mental health Counseling"\cite{bib21}, we have classified the topics of counseling questions into 8 categories, including background, cause, empathy, symptom, experience, cognition, behavior, and support. We consider a question to be more complex the more topics it contains, as illustrated in Figure 5.
For ease of description, we classify questions with 1-3 topics as simple, 4-6 topics as moderate, and 7 or more topics as complex. As shown in Table 5, as the complexity of the questions increases, the Empathy score from DSCs also continually improves. Similarly, the scores for Relevance, Helpfulness, and Empathy maintain a relatively high level.
Compared to the Zero-shot CoT prompting method, DSCs perform better in all the evaluation metrics, regardless of the complexity of the questions. This further illustrates the advantage of the DSCs model in handling problems of different complexities.
\begin{figure}[h]
\centering
\centerline{\includegraphics[scale=0.5]{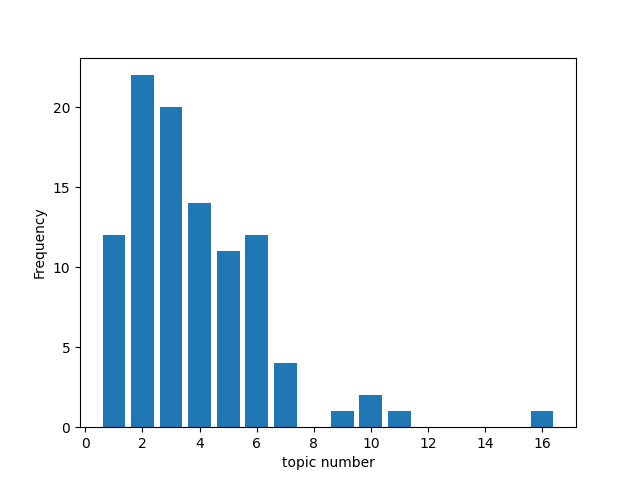}}%%图像路径：pic是文件夹名。
\caption{Distribution of Topic Quantities. The topics include eight categories: background, cause, empathy, symptom, experience, cognition, behavior, and support.}
\label{ov}
\end{figure}

\section{Discussions}
In this paper, we propose an approach for PLMs and LLMs to interact and complete the LTGM task. This approach addresses two main issues currently present in prompting methods: a lack of personalization (inability to change strategies based on the question) and a lack of comprehensive thinking. Specifically, we introduce a new method called Zero-shot DSC prompting, which incorporates Multiple DSCs to provide personalized strategies for individuals seeking help. Furthermore, we design a multiple-choice scheme to guide LLMs in considering a holistic strategy while answering questions, thereby providing optimal solutions. Extensive experiments have demonstrated the superiority and rationality of this model. In the future, further exploration is warranted to investigate the use of PLMs in generating efficient prompting to inspire LLMs to produce better results.

\section{Limitation}
This study has the following limitations. First, due to the lack of long mental health text datasets with mental health support strategies, researchers could only choose the PsyQA dataset for experimentation, which may restrict a comprehensive evaluation of the effects of mental health support strategy guidance. Additionally, there is a scarcity of manually annotated questions in PsyQA. To supplement training, we included machine-annotated questions from PsyQA, but as there are differences in the quality between machine and human annotations, this may impact the final results. Lastly, this study primarily focuses on the interaction methods between PLMs and LLMs, considering only GPT2 and GPT3.5-turbo, without considering other models. It is possible that other models may perform better in this regard.

\section{Ethical Considerations}
Finally, we discuss the potential ethical impacts of this work: The PsyQA dataset is a publicly available, well-established benchmark for emotional support conversation. The original providers have filtered sensitive information such as personally identifiable information (Sun et al., 2021). This work does not provide any treatment recommendations or diagnostic claims. It only provides mental health support guidance for LLMs. The decision to filter ethical risks is related to the use of LLMs. This work cannot replace human mental health counselors. In the event that help-seekers do not experience any improvement after interacting with a conversational system trained through multiple rounds of dialogue, we strongly recommend that they seek timely assistance from professional counselors or psychiatrists.

\bigskip
\noindent Thank you for reading these instructions carefully. We look forward to receiving your electronic files!

\bibliography{aaai24}

\begin{thebibliography}{21}
\providecommand{\natexlab}[1]{#1}

\bibitem[{Bosselut et~al.(2019)Bosselut, Rashkin, Sap, Malaviya, Celikyilmaz,
  and Choi}]{bib14}
Bosselut, A.; Rashkin, H.; Sap, M.; Malaviya, C.; Celikyilmaz, A.; and Choi, Y.
  2019.
\newblock {COMET:} Commonsense Transformers for Automatic Knowledge Graph
  Construction.
\newblock In \emph{{ACL} 2019, Volume 1: Long Papers}, 4762--4779.

\bibitem[{Brown et~al.(2020)Brown, Mann, Ryder, Subbiah, Kaplan, Dhariwal,
  Neelakantan, Shyam, Sastry, and Askell}]{bib5}
Brown, T.~B.; Mann, B.; Ryder, N.; Subbiah, M.; Kaplan, J.; Dhariwal, P.;
  Neelakantan, A.; Shyam, P.; Sastry, G.; and Askell, A. 2020.
\newblock Language Models are Few-Shot Learners.
\newblock In \emph{NeurIPS 2020, December 6-12, 2020, virtual}.

\bibitem[{Chowdhery et~al.(2022)Chowdhery, Narang, Devlin, Bosma, Mishra,
  Roberts, Barham, and Chung}]{bib6}
Chowdhery, A.; Narang, S.; Devlin, J.; Bosma, M.; Mishra, G.; Roberts, A.;
  Barham, P.; and Chung, H.~W. 2022.
\newblock PaLM: Scaling Language Modeling with Pathways.
\newblock \emph{ArXiv}, abs/2204.02311.

\bibitem[{Cohen(1960)}]{bib20}
Cohen, J. 1960.
\newblock A coefficient of agreement for nominal scales.
\newblock \emph{Educational and psychological measurement}, 20(1): 37--46.

\bibitem[{Hill(2009)}]{bib9}
Hill, C.~E., ed. 2009.
\newblock \emph{Helping skills: Facilitating, exploration, insight, and
  action}.
\newblock Reading, Mass.: American Psychological Association.

\bibitem[{Jiang(2012)}]{bib21}
Jiang, G. 2012.
\newblock \emph{The Theory and Practice of Psychological Counseling}.
\newblock Higher Education Press.

\bibitem[{Jr. et~al.(2013)Jr., Stiles, Bailer, and Hughes}]{bib4}
Jr., D. J.~R.; Stiles, W.~B.; Bailer, A.~J.; and Hughes, M.~R. 2013.
\newblock Impact of Exchanges and Client-Therapist Alliance in Online-Text
  Psychotherapy.
\newblock \emph{Cyberpsychology Behav. Soc. Netw.}, 16(5): 370--377.

\bibitem[{Li et~al.(2016)Li, Galley, Brockett, Gao, and Dolan}]{bib19}
Li, J.; Galley, M.; Brockett, C.; Gao, J.; and Dolan, B. 2016.
\newblock A Diversity-Promoting Objective Function for Neural Conversation
  Models.
\newblock In \emph{{NAACL} {HLT} 2016}, 110--119.

\bibitem[{Li et~al.(2022)Li, Li, Ren, Ren, and Chen}]{bib13}
Li, Q.; Li, P.; Ren, Z.; Ren, P.; and Chen, Z. 2022.
\newblock Knowledge Bridging for Empathetic Dialogue Generation.
\newblock In \emph{{AAAI} 2022}, 10993--11001.

\bibitem[{Liu et~al.(2021)Liu, Zheng, Demasi, Sabour, Li, Yu, Jiang, and
  Huang}]{bib10}
Liu, S.; Zheng, C.; Demasi, O.; Sabour, S.; Li, Y.; Yu, Z.; Jiang, Y.; and
  Huang, M. 2021.
\newblock Towards Emotional Support Dialog Systems.
\newblock In \emph{{ACL/IJCNLP} 2021, (Volume 1: Long Papers)}, 3469--3483.

\bibitem[{Papineni et~al.(2002)Papineni, Roukos, Ward, and Zhu}]{bib18}
Papineni, K.; Roukos, S.; Ward, T.; and Zhu, W. 2002.
\newblock Bleu: a Method for Automatic Evaluation of Machine Translation.
\newblock In \emph{Proceedings of the 40th Annual Meeting of the Association
  for Computational Linguistics}, 311--318.

\bibitem[{Peng et~al.(2023)Peng, Qin, Hu, Xie, and Li}]{bib11}
Peng, W.; Qin, Z.; Hu, Y.; Xie, Y.; and Li, Y. 2023.
\newblock {FADO:} Feedback-Aware Double COntrolling Network for Emotional
  Support Conversation.
\newblock \emph{Knowl. Based Syst.}, 264: 110340.

\bibitem[{Sabour et~al.(2023)Sabour, Zhang, Xiao, Zhang, Zheng, Wen, Zhao, and
  Huang}]{bib3}
Sabour, S.; Zhang, W.; Xiao, X.; Zhang, Y.; Zheng, Y.; Wen, J.; Zhao, J.; and
  Huang, M. 2023.
\newblock A chatbot for mental health support: exploring the impact of Emohaa
  on reducing mental distress in China.
\newblock \emph{Frontiers Digit. Health}, 5.

\bibitem[{Sun et~al.(2021)Sun, Lin, Zheng, Liu, and Huang}]{bib2}
Sun, H.; Lin, Z.; Zheng, C.; Liu, S.; and Huang, M. 2021.
\newblock PsyQA: {A} Chinese Dataset for Generating Long Counseling Text for
  Mental Health Support.
\newblock In \emph{{ACL/IJCNLP} 2021}, Findings of {ACL}, 1489--1503.

\bibitem[{Touvron et~al.(2023)Touvron, Lavril, Izacard, Martinet, and
  Lachaux}]{bib7}
Touvron, H.; Lavril, T.; Izacard, G.; Martinet, X.; and Lachaux, M. 2023.
\newblock LLaMA: Open and Efficient Foundation Language Models.
\newblock \emph{ArXiv}, abs/2302.13971.

\bibitem[{Tu et~al.(2022)Tu, Li, Cui, Wang, Wen, and Yan}]{bib12}
Tu, Q.; Li, Y.; Cui, J.; Wang, B.; Wen, J.; and Yan, R. 2022.
\newblock {MISC:} {A} Mixed Strategy-Aware Model integrating {COMET} for
  Emotional Support Conversation.
\newblock In \emph{{ACL} 2022}, 308--319.

\bibitem[{Wang et~al.(2023)Wang, Xu, Lan, Hu, Lan, Lee, and Lim}]{bib16}
Wang, L.; Xu, W.; Lan, Y.; Hu, Z.; Lan, Y.; Lee, R.~K.; and Lim, E. 2023.
\newblock Plan-and-Solve Prompting: Improving Zero-Shot Chain-of-Thought
  Reasoning by Large Language Models.
\newblock In \emph{{ACL} 2023}, 2609--2634.

\bibitem[{Wei et~al.(2022)Wei, Wang, Schuurmans, Bosma, Ichter, Xia, Chi, Le,
  and Zhou}]{bib15}
Wei, J.; Wang, X.; Schuurmans, D.; Bosma, M.; Ichter, B.; Xia, F.; Chi, E.~H.;
  Le, Q.~V.; and Zhou, D. 2022.
\newblock Chain-of-Thought Prompting Elicits Reasoning in Large Language
  Models.
\newblock In \emph{NeurIPS}.

\bibitem[{{World Health Organization}(2022)}]{bib1}
{World Health Organization}. 2022.
\newblock \emph{World mental health report: Transforming mental health for
  all}.

\bibitem[{Yao et~al.(2023)Yao, Yu, Zhao, Shafran, Griffiths, Cao, and
  Narasimhan}]{bib17}
Yao, S.; Yu, D.; Zhao, J.; Shafran, I.; Griffiths, T.~L.; Cao, Y.; and
  Narasimhan, K. 2023.
\newblock Tree of Thoughts: Deliberate Problem Solving with Large Language
  Models.
\newblock \emph{ArXiv}, abs/2305.10601.

\bibitem[{Zhang et~al.(2023)Zhang, Sheng, Zhou, and Chen}]{bib8}
Zhang, Z.~A.; Sheng, Y.; Zhou, T.; and Chen, T. 2023.
\newblock H\$\_2\$O: Heavy-Hitter Oracle for Efficient Generative Inference of
  Large Language Models.

\end{thebibliography}

% \section{Manual evaluation criteria}
\setcounter{secnumdepth}{2} 
\appendix
\section{Manual Evaluation Criteria}
\begin{table}[h]
    \centering
    \small
    \renewcommand{\arraystretch}{1.2} % 增加行高
    \begin{tabular}{p{1.5cm}p{6cm}}
    \hline
    Fluency & \\ \hline
    1 point & Numerous errors, difficult to understand \\
    2 points & Many errors, challenging to read \\
    3 points & Some errors, can be understood with effort \\
    4 points & Few errors, overall readable \\
    5 points & No errors, highly readable \\
    \hline
    Relevance & \\ \hline
    1 point & Completely irrelevant \\
    2 points & Low relevance, only touches on minor related content \\
    3 points & Partially relevant, key information missing \\
    4 points & Moderately relevant, covers most of the key content \\
    5 points & Completely relevant, provides accurate and comprehensive information \\
    \hline
    Helpfulness & \\ \hline
    1 point & No help or guidance \\
    2 points & Limited explanations/analysis/suggestions, offers minimal assistance \\
    3 points & Partially helpful in addressing key issues, but incomplete \\
    4 points & Provides some help, resolves most of the problems \\
    5 points & Highly helpful, offers clear and practical solutions \\
    \hline
    Empathy & \\ \hline
    1 point & No emotional response \\
    2 points & Slightly reflects emotions, but not evident or sincere \\
    3 points & Shows some warmth/sympathy/concern, room for improvement \\
    4 points & Displays a certain degree of warmth/sympathy/concern \\
    5 points & Filled with warmth/sympathy/concern, provides comfort and support \\
    \hline
    \end{tabular}
    \caption{Manual evaluation criteria }
    \label{table1}
\end{table}
\end{document}